\documentclass{article} 
\usepackage{iclr2021_conference,times}

\usepackage{amsmath,amsfonts,bm}

\def\eqref#1{equation~\ref{#1}}

\def\1{\bm{1}}

\DeclareMathAlphabet{\mathsfit}{\encodingdefault}{\sfdefault}{m}{sl}
\SetMathAlphabet{\mathsfit}{bold}{\encodingdefault}{\sfdefault}{bx}{n}

\usepackage{hyperref}
\usepackage{url}
\usepackage{graphicx}
\usepackage{booktabs}

\title{Deep Learning for Rheumatoid Arthritis:\\Joint Detection and Damage Scoring in X-rays}

\author{Krzysztof Maziarz\thanks{Correspondence to \texttt{krzysztof.maziarz@microsoft.com}}\\
Microsoft Research\\
\And
Anna Krason \\
University College London \\
\And
Zbigniew Wojna \\
Tensorflight \\
}

\iclrfinalcopy

\begin{document}

\nocite{*}

\maketitle

\begin{abstract}
Recent advancements in computer vision promise to automate medical image analysis. Rheumatoid arthritis is an autoimmune disease that would profit from computer-based diagnosis, as there are no direct markers known, and doctors have to rely on manual inspection of X-ray images. In this work, we present a multi-task deep learning model that simultaneously learns to localize joints on X-ray images and diagnose two kinds of joint damage: narrowing and erosion. Additionally, we propose a modification of label smoothing, which combines classification and regression cues into a single loss and achieves 5\% relative error reduction compared to standard loss functions. Our final model obtained 4th place in joint space narrowing and 5th place in joint erosion in the global RA2 DREAM challenge.
\end{abstract}

\section{Introduction}

Rheumatoid arthritis (RA) is an autoimmune disease which commonly affects joints in hands, wrists and feet, and can cause chronic pain~\citep{CDC:2020dg}. One of the impediments to its efficient diagnosis is the lack of direct markers for RA; rheumatologists have to rely on various clinical clues in order to determine whether a patient should start antirheumatic therapy~\citep{fukae2020convolutional}.

\begin{figure}[h]
    \centering
    \includegraphics[height=4.06cm]{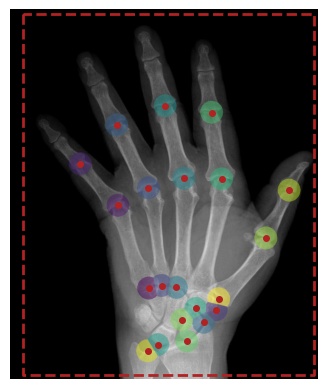}
    \includegraphics[height=4.06cm]{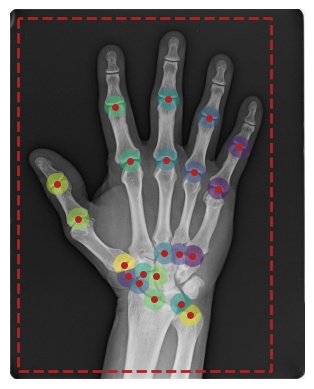}
    \includegraphics[height=4.06cm]{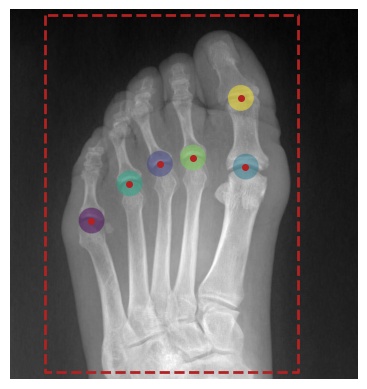}
    \includegraphics[height=4.06cm]{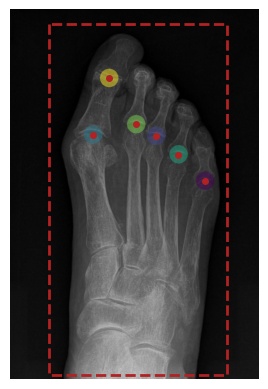}
    \caption{Four example images from the training dataset. The dashed box shows the region of interest computed during preprocessing, dots mark the ground-truth center of each joint, and colored regions correspond to the ground-truth segmentation mask for joint detection.}
    \label{fig:examples}
\end{figure}

Joint damage scoring is typically done by a highly-trained medical professional, who has to meticulously review several radiographic images of hands and feet. This procedure is not only costly and time-consuming, but also gives subjective results. The need for medical expertise may delay access to diagnosis; this is especially true in developing countries, where hospitals are often understaffed.

Recent advancements in deep learning promise to address these challenges, by \emph{learning} to assess joint damage directly from data. Early attempts include determining whether the patient has RA using various clinical data~\citep{fukae2020convolutional}, and diagnosing RA by training a convolutional neural network (CNN) on a radiographs dataset~\citep{ureten2020detection}.

One established metric to quantify joint damage is the Sharp/van der Heijde (SvH) method~\citep{van1995radiographic}. This scoring system separately evaluates two kinds of damage: joint space narrowing and joint erosion\footnote{For more background information about the two kinds of joint damage see Appendix~\ref{appendix:background}.}. For narrowing, 15 joints in each hand and 6 joints in each foot are given scores from 0 to 4. For erosion, 16 joints in each hand are given scores from 0 to 5, while 6 joints in each foot are scored from 0 to 10.
Finally, the overall SvH score is the sum of all narrowing and erosion scores, which is an integer ranging from 0 to 448.

In this paper, we propose a deep multi-task~\citep{caruana1997multitask} neural architecture that predicts joint narrowing and erosion scores following the SvH method. We make the following contributions:
\begin{itemize}
\item We design a deep CNN architecture to estimate SvH scores for RA damage. It simultaneously performs \emph{joint localization}, \emph{joint erosion assessment} and \emph{joint narrowing assessment}. 
\item We propose \emph{local label smoothing}, which includes class order in the cross entropy classification loss. With local label smoothing, we achieve a 5\% relative error reduction.  
\end{itemize}

\section{Related work}

There have been many approaches to automated RA diagnosis, both using deep CNNs, and classical machine learning techniques such as support vector machines (SVM). These works can be divided into those exploring a coarse-grained task of classifying patients as either normal or suffering from RA, and those that produce fine-grained SvH scores for all joints.

\paragraph{Detecting RA as binary classification}
Several works tried to predict whether a patient has RA by training a classification network on some patient data. \cite{fukae2020convolutional} used various clinical information converted into an image, which was then processed by an AlexNet~\citep{krizhevsky2012imagenet}. In contrast, other approaches used imaging data directly, either X-ray~\citep{ureten2020detection} or ultrasound~\citep{andersen2019neural}. However, modelling RA diagnosis with a simple binary variable lacks explainability, which is essential for clinical adoption~\citep{amann2020explainability}.

\paragraph{Detecting RA using the SvH method}
Building models that predict fine-grained SvH scores can lead to a much richer interplay between machine learning models and medical practitioners. Early works have explored this direction using simple machine learning models, such as SVMs~\citep{tashita2017automated} and shallow CNNs~\citep{hirano2019development}. Recently, modern CNN architectures~\citep{li2020ra2, israel2020ra2, pataki2020ra2, dimitrovsky2020multistage, stadler2020two, tran2020ra2} have been used to predict SvH scores as part of the RA2 Dream Challenge~\citep{ra2challenge}. As our method has also been submitted to this challenge, these results can be directly compared to ours, and they have been found to achieve similar results on the challenge test data. We differentiate ourselves from these methods in more detail in Section~\ref{sec:experiments}.

\section{Our method}

We approach SvH scoring of hand and foot images as simultaneous segmentation and classification. For training, we utilize four images per patient, with all narrowing and erosion scores annotated for each image. In theory, we could train a model to predict all scores from raw pixel data without localization cues. However, data in the medical domain is typically scarce, making this approach unfeasible, as it is not sample efficient. Therefore, we also assume access to annotations of center positions of all joints, and use that as an additional training signal.

\paragraph{Segmentation masks}

During data preprocessing, we convert joint center annotations into a segmentation mask, associating some pixels with the class tied to the corresponding joint, and some with an additional background class. As the optimal size of the relevant joint region is unclear, during the labelling stage we only annotate joint centers, and convert them to segmentation masks during preprocessing. In this way, we decouple generating the ground truth segmentation mask from annotating the data. 

We parametrize the conversion of joint centers into a mask with two hyperparameters $r$ and $R$, where $r \leq R$. First, for each pixel we find the closest joint center. If that closest center is at distance at most $r$, we use it as the ground truth class for that pixel. If this distance is more than $R$, we associate the given pixel with the background class. Pixels where the distance to the closest center falls into the $(r, R\rangle$ range we consider to lie on a boundary; we therefore \emph{do not assign them to any class}, and ignore them in the segmentation loss.

\paragraph{Training objective}\label{sec:objective}

We train our network to minimize a simple weighted average of three per-pixel objectives. First, for each pixel, we predict the corresponding joint class. In hands, 15 joints are relevant for narrowing and 16 joints for erosion; these two sets largely overlap apart from some joints in the wrist, yielding a total of 21 points of interest (Figure~\ref{fig:examples}, left). As there is a natural mapping between toes and fingers, we map the 6 feet joints (Figure~\ref{fig:examples}, right) into the shared set of 21 joint types. Together with the background class, the total number of classes for the joint localization head is 22. Next, we classify each positive (joint) pixel to predict a narrowing score (5 possible values) and an erosion score (6 values). Intuitively, this multi-task formulation decouples joint localization from damage assessment, as narrowing and erosion is detected with the same network head, irrespective of the underlying joint. All three prediction tasks are modelled as classification.

As erosion in feet is scored on a 0-10 scale, we train the model to match \emph{half} of the actual score, and then multiply the predictions by 2 during inference. In that way, the range of values for erosion scores is aligned across all joints; this is crucial as damage prediction is joint-agnostic.

\paragraph{Local label smoothing}

As explained in the previous section, we model the prediction of narrowing and erosion scores as a classification task. Casting the prediction of a small integer as classification instead of (constrained) regression is common practice in many machine learning models, as such approach is typically robust and easy to train~\citep{kozakowski2019forecasting,wojna2020holistic}. In preliminary experiments, we found that training for classification worked significantly better than regression. However, using classification ignores the inherent ordering of the classes: predicting a narrowing score of 1 is a much better answer if the ground truth is 0, than if the ground truth is 4.

In this work, since SvH scores are small integers, we propose to model the task as classification, but inject a small regression-inspired bias into the ground truth label. Similarly to label smoothing~\citep{muller2019does}, we use a smoothed one-hot vector as ground truth, placing most of the probability mass on the target class, and a small amount of mass on the other classes. However, in contrast to classical label smoothing, we only move probability mass to neighbouring classes: if the ground truth label is $x$, we place extra probability mass on classes $x-1$ and $x+1$ ($\frac{p}{2}$ each), and the remaining $1 - p$ on $x$. Note that $x$ can be one of the boundary classes (either $0$ or the maximum possible score), in which case $1 - \frac{p}{2}$ of probability mass is placed on the ground truth class $x$.

\paragraph{Model}

We utilise the U-Net architecture, which was shown to achieve strong performance on biomedical image segmentation~\citep{ronneberger2015u, cciccek20163d, schlemper2019attention}.
As the encoder, we use the EfficientNet B5 network~\citep{tan2019efficientnet}; for the decoder we use traditional upsampling convolutional layers~\citep{zeiler2011adaptive}. Following common practice for U-Nets, we add skip connections from encoder layers into decoder layers at a matching resolution. As the pixel location within the image can aid the classification into joint types, we use the CoordConv technique~\citep{liu2018intriguing}, and concatenate the input image with two additional channels, corresponding to x and y pixel coordinates.

The final feature map from the last decoder layer is used as input to three separate per-pixel classifiers, which give the output for each of the three tasks specified earlier. Therefore, our architecture supports multiple tasks via the \emph{shared bottom} paradigm, as most of the network is shared. This works well since the tasks are highly related; however, more complex methods could explicitly trade off positive and negative transfer~\citep{ma2019snr, maziarz2019flexible, zhang2020overcoming}.

During inference we produce the final narrowing and erosion scores as a weighted average, using the softmax predictions produced by the model as weights.

\section{Experiments}\label{sec:experiments}

\paragraph{Training data}

We obtained the training data from the RA2 DREAM Challenge~\citep{sun2021crowdsourcing,bridges2010radiographic,ormseth2015effect}, which contains data from 367 patients, with four images per patient (both hands and feet). We split the provided training data into train and validation sets by using the first out of eight folds as the validation set and the rest of the data as the training set. We show four example images from the training set in Figure~\ref{fig:examples}.

The input images come with a variable amount of irrelevant black pixels at the borders, while only the pixels that correspond to the hand or foot are informative. To further normalize the data, we used classical computer vision techniques such as the canny algorithm to detect the bounding box of the hand or foot visible in the image. We then cropped each image to the relevant bounding box, and scaled the resulting image to a resolution of 864 x 928. In Figure~\ref{fig:examples}, we mark the bounding box computed for the example images with dashed rectangles. As the amount of data was limited, we also employed several data augmentation techniques: rotation, scaling and horizontal flips.

\paragraph{Results}

We trained the network using the AdamW optimizer~\citep{loshchilov2017decoupled} and OneCycle learning rate scheduling method~\citep{smith2017cyclical}. We tuned all hyperparameters by performing small-scale experiments on our local validation set. We compared experiments by computing the root mean squared error (RMSE) of SvH score predictions for all joints, averaged over all examples in the validation set.

\begin{figure}[t]
    \centering
    \includegraphics[width=0.49\columnwidth]{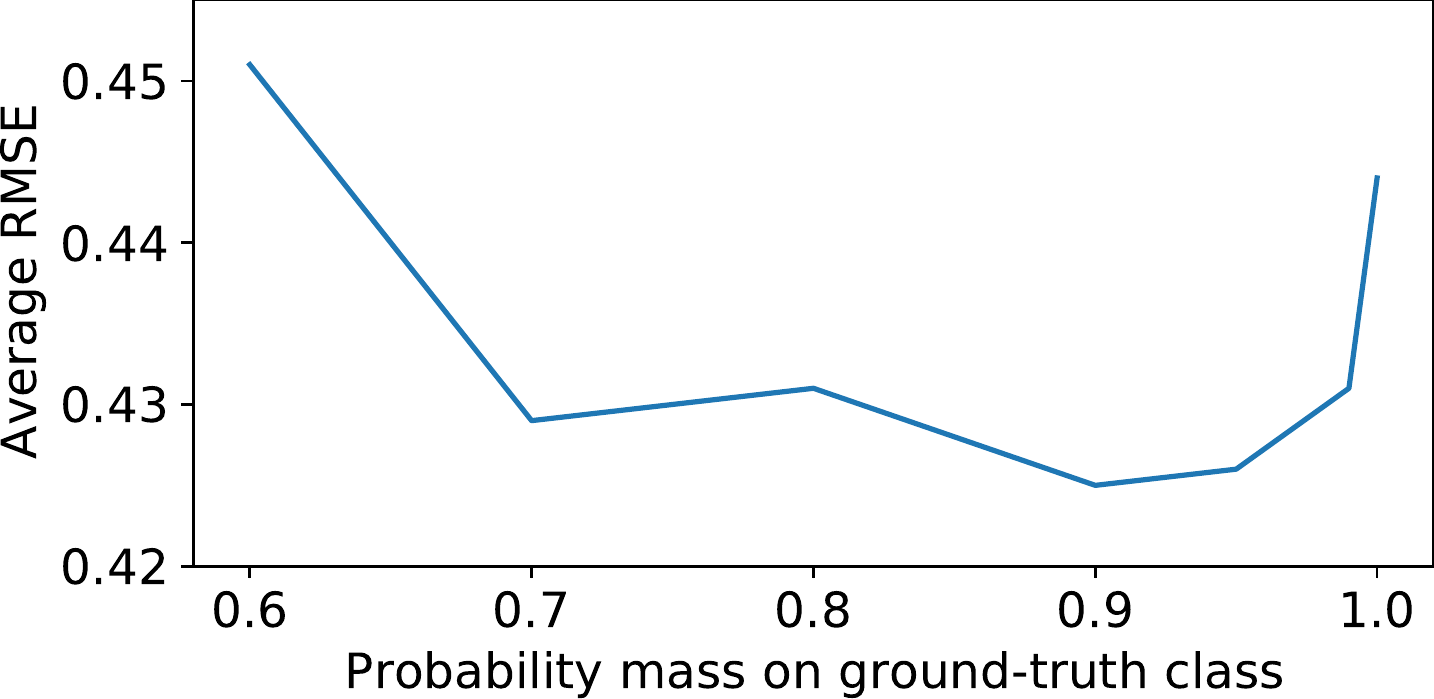}
    \includegraphics[width=0.49\columnwidth]{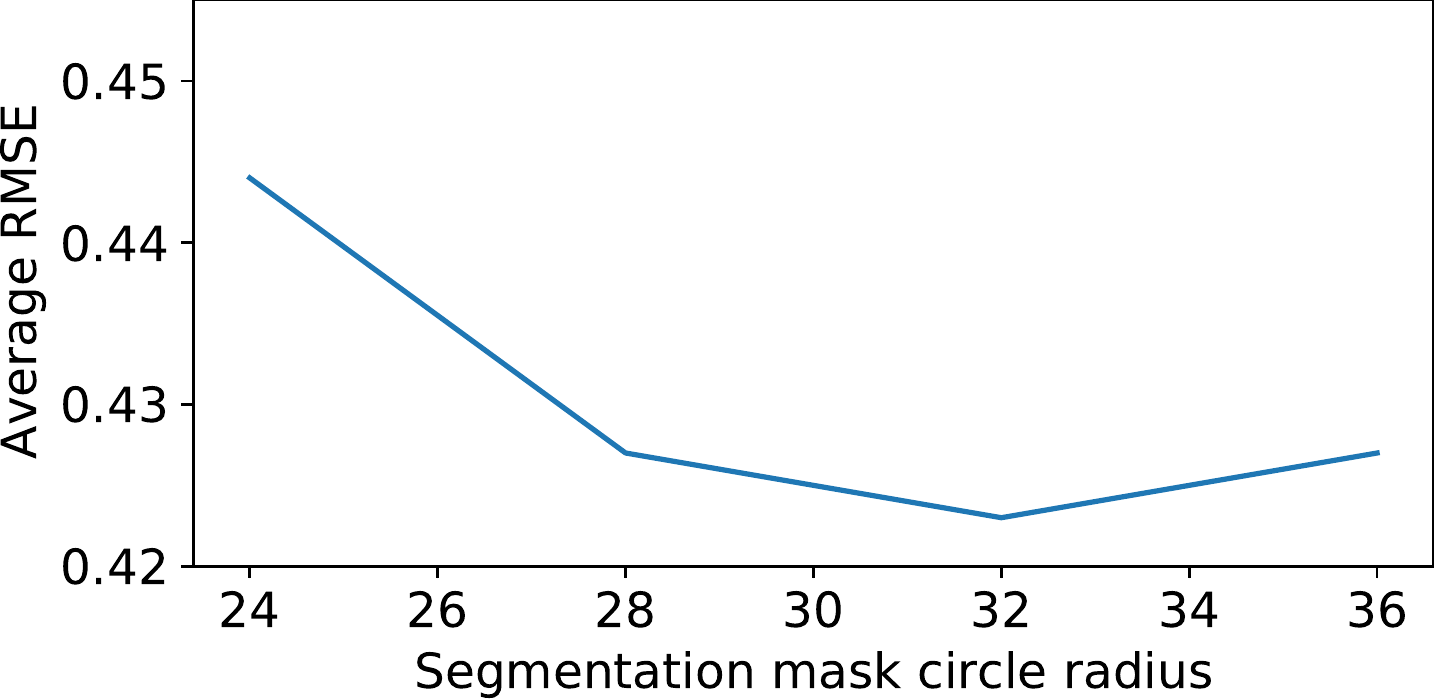}
    \caption{The effect of varying local label smoothing (left) and segmentation circle size (right) on downstream model performance.}
    \label{fig:ablation}
\end{figure}

After hyperparameter tuning, we used a well performing setting to train our model 8 times on all available data. We used an ensemble of these 8 models to predict SvH scores on unseen test data, resulting in mean RMSE of 0.4075 for narrowing, and 0.4607 for erosion. On the final leaderboard, these values fall within 10\% relative difference of the top result, obtaining 4th and 5th place for narrowing and erosion, respectively. Small gains that other models submitted to the challenge achieved on top of our results~\citep{li2020ra2, israel2020ra2, pataki2020ra2} can be attributed to using much more complex pipelines (with up to three separate steps, each utilizing a different ML model), and sometimes performing extensive human labour (by authors learning the basics of RA scoring, and then manually re-annotating the data to obtain more labels). In contrast, we show that a simple architecture trained in a multi-task setting can obtain results competitive with the state-of-the-art. See Appendix~\ref{appendix:challenge} for a detailed comparison of top performing models in the challenge.

\paragraph{Ablation study}

To better understand which hyperparameters are important to achieve good performance, we performed an ablation study on the validation set. While many hyperparameters had limited impact on downstream performance or were easy to tune, we found two that showed interesting trends: the degree of local label smoothing, and the radius $r$ used to compute joint segmentation masks.

We show the ablation results in Figure~\ref{fig:ablation}. We see that shifting $0.1$ of the probability mass to classes adjacent to the ground truth improves results, yielding a relative error rate reduction of approximately $5\%$. Moreover, we see that to achieve optimal performance $r$ has to be tuned rather carefully, with the optimal value around $r = 32$ (given a fixed value of $R = 40$).
 
\section{Conclusion}

In this work, we have shown a deep neural model which simultaneously localizes joints and assesses the severity of rheumatoid arthritis. This was enabled by posing both objectives as pixel-level classification tasks, and training the model in a multi-task setting. Moreover, we proposed local label smoothing, which allows to smoothly interpolate between a classification and a regression objective.

\clearpage

\bibliography{references}
\bibliographystyle{iclr2021_conference}

\clearpage

\appendix

\section{Background on joint damage assessment}\label{appendix:background}

Rheumatoid arthritis is typically associated with two kinds of joint damage: \emph{joint space narrowing} and \emph{bone erosion}. Narrowing affects the joint cartilage, which ensures that the distance between interacting bones allows for a good range of motion. Narrowing means that the cartilage can no longer maintain a healthy distance between the bones, which leads to increased pressure and friction. Erosion is related to \emph{bone resorption}, which is a process in which the body breaks down bone tissue, releasing the minerals into the bloodstream. While normally this is a part of a healthy life cycle of bone tissue, it occurs excessively in patients suffering from rheumatoid arthritis, leading to irreversible damage.

\section{Comparison of top submissions to the RA2 DREAM challenge}\label{appendix:challenge}

The organizers of the RA2 DREAM challenge report full results of the best performing submissions in \citet{sun2021crowdsourcing}. Carefully comparing the approaches, we see that all competing methods use pipelines containing at least two steps, which typically are joint detection followed by extracting image patches and separately scoring them. Some also include a third stage, which is either post-processing to combine all results for a single patient, or pre-processing to normalize the orientation of the input images. Additionally, \cite{li2020ra2} "served as an extra newbie radiologist to manually score all the training data", therefore making use of more labels than other approaches. In contrast, our method only needs a single model trained in a multi-task setting, and did not use extra labels, while still attaining performance close to the state-of-the-art.

\end{document}